%% file: main.tex
\documentclass[10pt,twocolumn,letterpaper]{article}

\usepackage{cvpr}              

\usepackage[ruled]{algorithm2e}
\usepackage{makecell}
\usepackage{multirow}
\usepackage{graphicx} 
\usepackage{threeparttable} 


\definecolor{cvprblue}{rgb}{0.21,0.49,0.74}
\usepackage[pagebackref,breaklinks,colorlinks,allcolors=cvprblue]{hyperref}
\usepackage[accsupp]{axessibility}
\def\confName{CVPR}
\def\confYear{2026}

\title{HyperFM: An Efficient Hyperspectral Foundation Model with Spectral Grouping}

\author{
Zahid Hassan Tushar, Sanjay Purushotham\\
University of Maryland, Baltimore County\\
Baltimore, MD, USA\\
{\tt\small ztushar1@umbc.edu, psanjay@umbc.edu}
}

\begin{document}
\maketitle
\input{MySections/abstract}
\input{MySections/introduction}
\input{MySections/related}

\input{MySections/Data}
\input{MySections/Approach}

\input{MySections/Experiments}
\input{MySections/Results}
\input{MySections/Discussions}
\input{MySections/Conclusions}
\input{MySections/ack}

{
    \small
    \bibliographystyle{ieeenat_fullname}
    \bibliography{main}
}

\input{MySections/suppl}


\end{document}

%% file: MySections/abstract.tex
\begin{abstract}
The NASA PACE mission provides unprecedented hyperspectral observations of ocean color, aerosols, and clouds, offering new insights into how these components interact and influence Earth’s climate and air quality. Its Ocean Color Instrument measures light across hundreds of finely spaced wavelength bands, enabling detailed characterization of features such as phytoplankton composition, aerosol properties, and cloud microphysics. However, hyperspectral data of this scale is large, complex, and difficult to label, requiring specialized processing and analysis techniques. Existing foundation models, which have transformed computer vision and natural language processing, are generally trained on standard RGB imagery and therefore struggle to interpret the continuous spectral signatures captured by PACE. 
While recent advances have introduced hyperspectral foundation models, they are typically trained on cloud-free observations and often remain limited to single-sensor datasets due to spectral inconsistencies across instruments. Moreover, existing models tend to be parameter-heavy and computationally expensive, limiting scalability and adoption in operational settings. To address these challenges, we introduce \textbf{HyperFM}, a parameter-efficient hyperspectral foundation model that leverages intra-group and inter-group spectral attention along with hybrid parameter decomposition to better capture spectral spatial relationships while reducing computational cost. \textbf{HyperFM} demonstrates consistent performance improvements over existing hyperspectral foundation models and task-specific state-of-the-art methods across four benchmark downstream atmospheric cloud property retrieval tasks. To support further research, we additionally release \textbf{HyperFM250K}, a large-scale hyperspectral dataset from the PACE mission that includes both clear and cloudy scenes. Code and data are publicly available on \url{https://github.com/umbc-sanjaylab/HyperFM}.
\end{abstract}

%% file: MySections/introduction.tex
\section{Introduction}
Clouds play a central role in the Earth’s climate system~\cite{okamura2017feasibility}. They regulate the energy budget by scattering and absorbing solar and terrestrial radiation and strongly influence the hydrological cycle. Even small perturbations in cloud properties can alter radiative forcing and climate feedbacks, yet these properties remain one of the largest sources of uncertainty in climate projections~\cite{ipcc_website}. Thus, accurate retrievals of cloud optical and microphysical properties such as cloud optical thickness (COT), cloud effective radius (CER), cloud water path (CWP), and cloud top height (CTH) are essential for improving climate models, weather forecasting, and atmospheric process understanding~\cite{huang2024optimal}. Satellite radiances are typically used to estimate these quantities~\cite{nakajima1990determination,nataraja2022segmentation,li2023transfer}, often through 1D look-up-tables and optimal inversion algorithms, which are computationally intensive and can introduce systematic biases~\cite{nakajima1990determination}. Recent deep learning approaches have shown improvements by exploiting spatial context~\cite{tushar2024cloudunet,tushar2025joint}, and some works incorporate auxiliary variables to boost accuracy~\cite{li2024physics}, though such designs can limit cross-sensor generalization. This motivates the development of a robust, transferable feature extractor analogous to foundation models in computer vision.

Foundation models such as CLIP~\cite{radford2021learning} and DALL-E~\cite{ramesh2021zero} learn general-purpose visual representations from large, diverse datasets and can be adapted to numerous downstream tasks with minimal labeled data~\cite{dosovitskiy2020image}. Similar trends are emerging in remote sensing~\cite{liu2024remoteclip}, healthcare~\cite{he2024foundation}, and other scientific domains~\cite{yang2024large}. In remote sensing, foundation models now span RGB~\cite{christie2018functional}, SAR~\cite{zhao2020opensarurban}, multispectral~\cite{sumbul2019bigearthnet,helber2019eurosat}, and hyperspectral imagery~\cite{wang2025hypersigma,braham2025spectralearth}. Most progress, however, has concentrated on RGB and multispectral data, where large curated datasets enable scale~\cite{cong2022satmae,noman2024rethinking}. Hyperspectral foundation models have only recently emerged~\cite{wang2025hypersigma,braham2025spectralearth}. Because no large-scale HSI datasets existed, both \textit{HyperSigma}~\cite{wang2025hypersigma} and \textit{SpectralEarth}~\cite{braham2025spectralearth} constructed new corpora and achieved competitive performance on small HSI benchmarks. However, both datasets deliberately exclude cloudy scenes (less than ten percent cloud cover), making them unsuitable for cloud-property retrieval, the focus of this work. This gap highlights the need for a large-scale, curated hyperspectral dataset explicitly designed for atmospheric cloud-focused foundation model development.

Our contributions are summarized as follows:
\begin{itemize}
    \item A large, AI-ready hyperspectral dataset, called \textbf{HyperFM250k}, derived from NASA PACE-Ocean Color Instrument(OCI) observations. \textbf{HyperFM250k} is the first dataset to provide full-spectrum radiance measurements over both cloudy and non-cloudy regions at scale, enabling benchmarking of foundation models for pixel-wise regression tasks involving cloud properties.
    \item A new parameter-efficient hyperspectral foundation model, \textbf{HyperFM}. It introduces a \textit{Group Embed module} and \textit{Hypoformer} blocks to jointly capture spectral and spatial structure while maintaining computational efficiency.
        \item Benchmarking on four cloud-property retrieval tasks, demonstrating that \textbf{HyperFM} outperforms state-of-the-art deep learning models and existing hyperspectral foundation models by more than 32\% while using roughly 50\% fewer parameters.
\end{itemize}

%% file: MySections/related.tex
\section{Related Works}
\label{sec:related}

Estimating cloud optical and microphysical properties such as cloud optical thickness, cloud effective radius, cloud water path, and cloud top height, relies on measurements from either active sensors (e.g., radar, lidar~\cite{winker2003calipso}) or passive instruments (e.g., satellite radiance observations~\cite{lindsey2002modis,nasa_pace_oci_2024}). While active sensors provide high-accuracy measurements, they are costly and offer limited spatial coverage. Passive sensors, in contrast, deliver continuous global observations, making them the primary data source for large-scale retrieval. Traditionally, these properties are derived using physics-based 1D lookup tables and optimal inversion algorithms~\cite{nakajima1990determination,poulsen2012cloud}, which are computationally demanding and memory intensive. Recently, AI and machine learning approaches have emerged as alternatives for cloud properties retrievals~\cite{okamura2017feasibility}. Most methods adopt U-Net–style architectures, including \textit{UNet}~\cite{nataraja2022segmentation,li2023transfer}, \textit{CloudUNet}~\cite{tushar2024cloudunet}, and \textit{CAM}~\cite{tushar2025joint}, which are trained to estimate one or multiple cloud parameters within a single framework. However, these methods are often task-specific, rely on a limited subset of radiance bands (typically 2–8), and frequently depend on auxiliary data to reduce uncertainty. Moreover, many of these models are trained on private datasets, restricting reproducibility and benchmarking. 

Recent advances in foundation models~\cite{awais2025foundation} have extended to the remote sensing domain, offering a unified way to transfer knowledge from large-scale datasets to diverse downstream tasks~\cite{xiao2025foundation}. Models pre-trained on massive multispectral and RGB imagery have demonstrated strong adaptability for tasks such as change detection, semantic segmentation, and land-cover classification~\cite{cong2022satmae,guo2024skysense,noman2024rethinking,li2024s2mae}. Both convolution based (such as ResNet~\cite{he2016deep}) and transformer based architectures (such as Vision Transformer (ViT)~\cite{dosovitskiy2020image} and Swin Transformer~\cite{liu2021swin}) have been explored, together with self supervised objectives such as contrastive learning~\cite{stojnic2021self,akiva2022self} and masked auto encoding~\cite{reed2023scale}. While RGB pretrained models remain restricted to RGB data, multispectral models must handle varying band configurations and therefore require additional adaptation at the embedding stage.

Hyperspectral foundation models extend this line of work to inputs containing hundreds of spectral channels. HyperSigma~\cite{wang2025hypersigma} uses separate spatial and spectral modules, whereas SpectralEarth and HyperFree~\cite{li2025hyperfree,braham2025spectralearth} employ projection networks to reduce spectral dimensionality before representation learning. These models are trained on large cloud free datasets and achieve competitive performance on several downstream tasks, but their performance decreases on applications that depend on cloudy observations because such pixels are absent during pretraining. Further progress is constrained by the lack of hyperspectral datasets with substantial cloud coverage, which motivates our HyperFM250k dataset with approximately sixty percent cloud presence.

Parameter efficient modeling has been explored through post training and pre training compression. Post training approaches include model compression~\cite{noach2020compressing} and parameter efficient fine tuning such as adapters~\cite{li2025hyperfree,zhuang2025argus} and LoRA~\cite{zanella2024low}. Pre training approaches modify architectural components and include low rank factorization~\cite{hrinchuk2020tensorized,liu2021enabling}, hybrid factorization~\cite{thakker2020rank}, and tensor train decomposition~\cite{li2022hypoformer}. Although these methods reduce parameters, they may restrict expressivity or retain quadratic attention cost.
Recently, hybrid tensor-train approaches~\cite{li2022hypoformer} have been proposed that combine linear layers with tensor-train decomposition to maintain full-rank expressivity while improving efficiency. We adopt this technique in our HyperFM model to balance representational power and computational scalability.

%% file: MySections/Data.tex
\begin{figure*}[tb]
    \centering
    \includegraphics[width=0.95\linewidth]{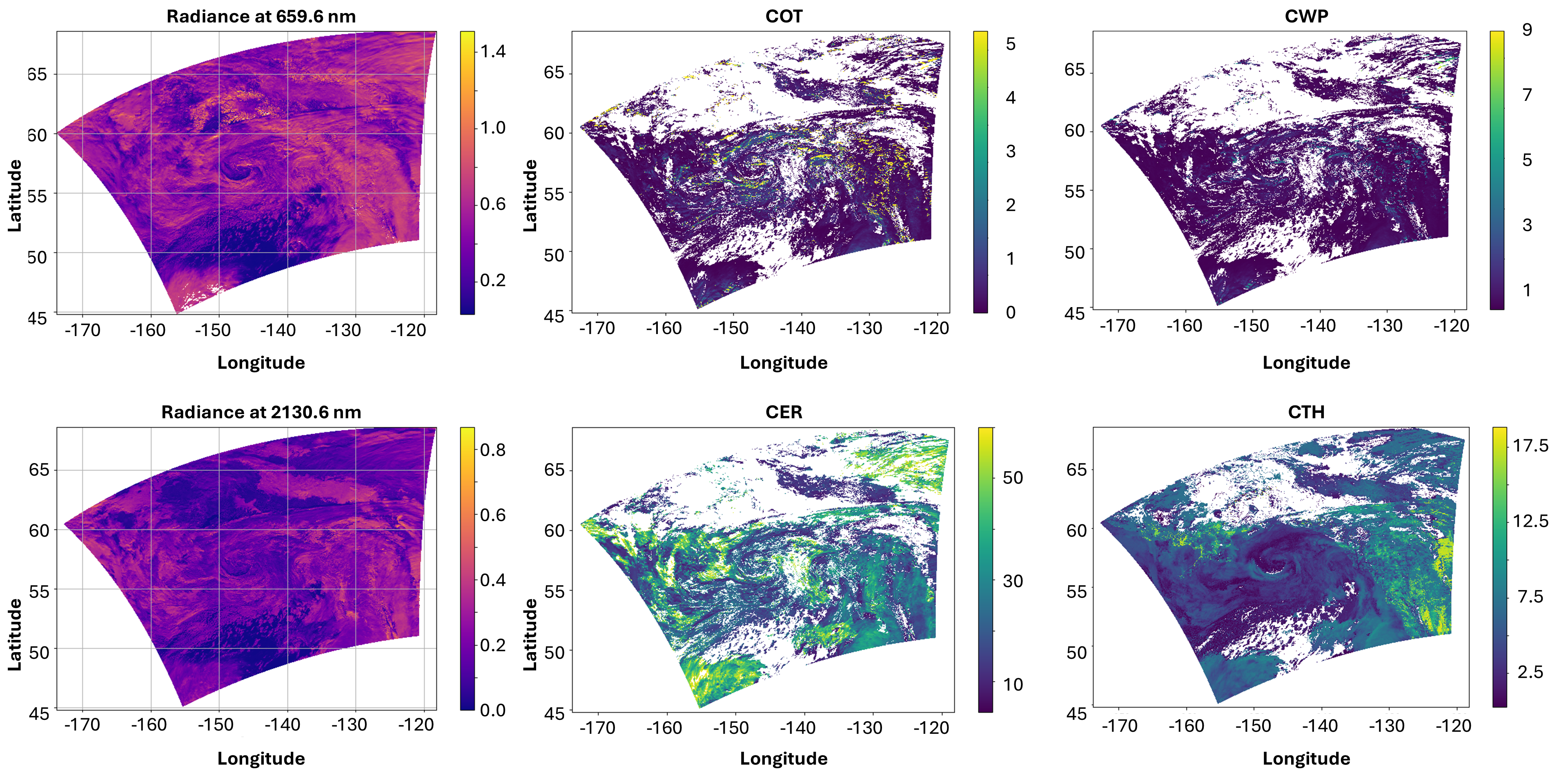}
    \caption{\textbf{Left Column}: PACE Level 1B radiance observations at 659.6nm (top) and 2130.6nm (bottom); \textbf{Middle Column}: PACE Level 2 products COT (top) and CER (bottom); \textbf{Right Column}: PACE Level 2 products CWP (top) and CTH (bottom).}
    \label{fig:samples}
\end{figure*}

\section{Hyperspectral Data}
\label{sec:data}

Hyperspectral imaging from space provides detailed information about the Earth’s surface and atmosphere, and recent missions have greatly expanded data availability. Key examples include the Environmental Mapping and Analysis Program (EnMAP)~\cite{kaufmann2012science}, the Hyperspectral Precursor and Application Mission (PRISMA)~\cite{pignatti2013prisma}, and the upcoming Copernicus Hyperspectral Imaging Mission for the Environment (CHIME)~\cite{nieke2018towards}. These missions primarily support land oriented HSI analysis and are designed for high resolution terrestrial applications such as vegetation mapping, mineralogy, and environmental monitoring. However, they are less suitable for ocean, polar, or atmospheric studies, including cloud and aerosol optical or microphysical property retrievals, which are essential for climate analysis and weather forecasting.

NASA’s Plankton, Aerosol, Cloud, ocean Ecosystem (PACE) mission, which includes the Ocean Color Instrument (OCI) and the HARP-2 polarimeter, differs from these land focused HSI sensors while sharing core hyperspectral characteristics. OCI provides contiguous VNIR hyperspectral sampling with high radiometric quality but is optimized for ocean color and atmospheric retrievals. It offers dense spectral coverage from 350 to 890 nm and multispectral SWIR channels at a much coarser spatial resolution of roughly 1.2 km, whereas land missions provide full VNIR–SWIR hyperspectral coverage at 30 m ground sampling distance. OCI also provides daily global observations, enabling large scale temporal pretraining but limiting fine spatial interpretation. These distinctions make OCI complementary to traditional hyperspectral missions and well aligned with cloud, aerosol, and ocean science applications. Earlier multispectral sensors such as MODIS~\cite{platnick2015modis} and VIIRS~\cite{platnick2019viirs} have supported cloud and aerosol retrievals, but their limited spectral coverage and data quality constraints reduce their suitability for accurate atmospheric microphysical inference compared with OCI. 

\begin{table}[tb]
\centering
\small
\caption{Comparison of Hyperspectral Imaging Missions}
\label{tab:hsi_mission_comparison}
\resizebox{\linewidth}{!}{
\begin{tabular}{lcccc}
\toprule
\textbf{Characteristic} & \textbf{PACE} & \textbf{EnMAP} & \textbf{PRISMA} & \textbf{CHIME} \\
\midrule
Mission            & Ocean--atmosphere            & Land            & Land                & Land \\
Spectral Range     & 350--890 nm (HSI) + SWIR MS  & 420--2450 nm    & 400--2500 nm        & 400--2500 nm \\
Spatial Resolution & {1.2 km}                & 30 m            & 30 m                & $\sim$30 m \\
Coverage           & {Daily global}        & Regional/global & Tasked              & Global \\
Target application           & Ocean color, Cloud, aerosol properties  & Land mapping    & Land mapping  & Land mapping \\
\bottomrule
\end{tabular}
}
\end{table}
While hyperspectral foundation models have been pretrained on land oriented HSI datasets such as SpectralEarth~\cite{braham2025spectralearth}, no large scale hyperspectral dataset currently exists for training foundation models tailored to atmospheric science applications. As summarized in~\cref{tab:hsi_mission_comparison}, land focused HSI datasets are typically acquired over non cloudy terrestrial regions and therefore lack the spectral signatures and variability associated with cloudy conditions. As a result, they are not suitable for finetuning models intended for cloud, aerosol, or atmospheric microphysics retrieval, since they do not capture the patterns contained in cloudy pixels across the hyperspectral bands. This limitation highlights the need for new datasets that support both pretraining and finetuning of hyperspectral foundation models for atmospheric science. To address this gap, we introduce \textbf{HyperFM250K}, a large scale hyperspectral dataset derived from the NASA PACE mission, containing 2262 granules collected across entire global. Below, we describe the data acquisition and preprocessing steps used to make this dataset AI ready for foundation model training.

\paragraph{Data acquisition:} HyperFM250K data were obtained from NASA’s EarthData portal using an automated Python based retrieval script. We downloaded 2262 Level 1B granules spanning May 2024 to April 2025. Level 1B data include radiometric calibration, instrument corrections, and geolocation processing, converting raw sensor measurements into physically meaningful radiance values. For downstream evaluation, we also collected the corresponding Level 2 cloud products, including cloud optical thickness (COT), cloud effective radius (CER), cloud water path (CWP), and cloud top height (CTH), using the same retrieval workflow. Each granule has spatially varying dimensions determined by the sensor’s geolocation mapping, with each pixel corresponding to approximately 1.2 km on the Earth’s surface. The Level 1B radiance measurements contain 291 spectral channels: the blue spectrometer spans $314\sim 605 nm$, the red spectrometer spans $600\sim 895nm$, and the SWIR spectrometer spans $940\sim 2258nm$. A sample is shown in~\cref{fig:samples}.

Our preprocessing pipeline, summarized in \cref{alg:one}, converts PACE–OCI Level 1B and Level 2 granules into valid hyperspectral patches through four steps. First, Level 1B granules and their quality flag arrays are loaded; pixels marked as invalid are replaced with NaN values, ensuring that subsequent processing ignores corrupted measurements. The same masking procedure is applied to Level 2 granules once a temporal match is found. Second, each Level 1B granule is paired with a Level 2 granule whose timestamp differs by no more than ten seconds, providing spatially and temporally aligned observations of the same scene. Third, a sliding window of size $96\times96$ is applied to the aligned granules to extract hyperspectral patches suitable for downstream analysis. Finally, patches containing more than one percent NaN values are discarded, yielding a curated set of high quality patches for training and evaluation.
After preprocessing, we obtain a large dataset of clean, spatially and temporally consistent hyperspectral patches suitable for foundation model pretraining and downstream analysis. A comparison of our dataset with existing hyperspectral datasets is provided in~\cref{tab:data_summary}.

\begin{table*}[t]
\centering
\small
\caption{Summary of hyperspectral datasets. HyperFM250k provides entire global coverage, high cloud coverage, suitable for atmospheric cloud property retrievals.}
\label{tab:data_summary}
\resizebox{\textwidth}{!}{
\begin{tabular}{lcccccccc}
\toprule
\textbf{Dataset / Method} & \textbf{Dates} & \textbf{\#Samples} & \textbf{Patch Size} & 
\textbf{Sensor(s)} & \textbf{Bands} & \textbf{GSD} & \textbf{Region Coverage: Land, Ocean, Polar} & \textbf{Cloud Coverage} \\
\midrule
HySpecNet-11k~\cite{fuchs2023hyspecnet} 
& -- & $\sim$11k & $128\times128$ & EnMAP & 202 & 30m & Yes, No, No & $<10\%$ \\

MSST~\cite{scheibenreif2023masked} 
& -- & $\sim$20k & $64\times64$ & EnMAP & 224 & 30m & Yes, No, No & $0\%$ \\

HSIHybrid~\cite{wang2024hsimae} 
& -- & $\sim$4m & $9\times9$ & Multiple & Variable & Variable & Yes, No, No & $<5\%$ \\

HyperGlobal-450k~\cite{wang2025hypersigma} 
& 2011--2017 & $\sim$447k & $64\times64$ & EO-1 and Gaofen-5B & 242 and 330 & 30m & Yes, No, No & $<5\%$ \\

SpectralEarth~\cite{braham2025spectralearth} 
& 2022--04 to 2024--04 & $\sim$538k & $128\times128$ & EnMAP & 202 & 30m & Yes, No, No & $<10\%$ \\

\textbf{HyperFM250k (ours)} 
& 2024--05 to 2025--04 & $\sim$250k & $96\times96$ & PACE OCI & 291 & 1.2km & Yes, Yes, Yes & $\mathbf{>60\%}$ \\
\bottomrule
\end{tabular}
}
\end{table*}






\begin{algorithm}[t]
\caption{Preprocessing Steps for HyperFM250K dataset}
\label{alg:one}

\KwIn{Level 1B granules $G_{1B}$, Level 2 granules $G_{2}$, patch size $(96,96)$, NaN threshold $T_{\mathrm{nan}}=0.01$, time tolerance $\Delta t=10s$}

\KwOut{Set of valid hyperspectral patches}

\ForEach{granule $g_{1B} \in G_{1B}$}{
    \tcp{Step 1: Mask invalid pixels in Level 1B}
    Load $D_{1B}$ and quality flags $Q_{1B}$\;
    
    Set $D_{1B}[Q_{1B}=1] \gets \mathrm{NaN}$\;

    \tcp{Step 2: Find matching Level 2 granule}
    \ForEach{$g_{2} \in G_{2}$}{
        \If{$|\mathrm{timestamp}(g_{1B})-\mathrm{timestamp}(g_{2})| \leq \Delta t$}{
            Load $D_{2}$ and quality flags $Q_{2}$\;
            
            Set $D_{2}[Q_{2}\neq0] \gets \mathrm{NaN}$\;
            
            Align $D_{1B}$ and $D_{2}$\;
        }
    }

    \tcp{Step 3: Extract patches}
    Slide a $(96,96)$ window over aligned data\;

    \ForEach{patch $P$}{
        Compute $\mathrm{NaN\_fraction}$\;
        
        \If{$\mathrm{NaN\_fraction} < T_{\mathrm{nan}}$}{
            Store patch $P$\;
        }
    }
}
\end{algorithm}

%% file: MySections/Approach.tex
\section{Our HyperFM Foundation Model}

We introduce \textbf{HyperFM}, a hyperspectral foundation model designed to learn rich and globally consistent representations from large scale hyperspectral datasets with extensive cloud coverage. \textbf{HyperFM} combines a new \textbf{\textit{`Group Embed'}} module with \textbf{\textit{Hypoformer}} blocks~\cite{li2022hypoformer} to efficiently model both spectral and spatial relationships while remaining parameter efficient. These architectural components address the high dimensional nature of hyperspectral satellite data by providing compact yet expressive representations through local and global attention and efficient tensor train based parameterization.

\begin{figure}[tb]
    \centering
    \includegraphics[width=\linewidth]{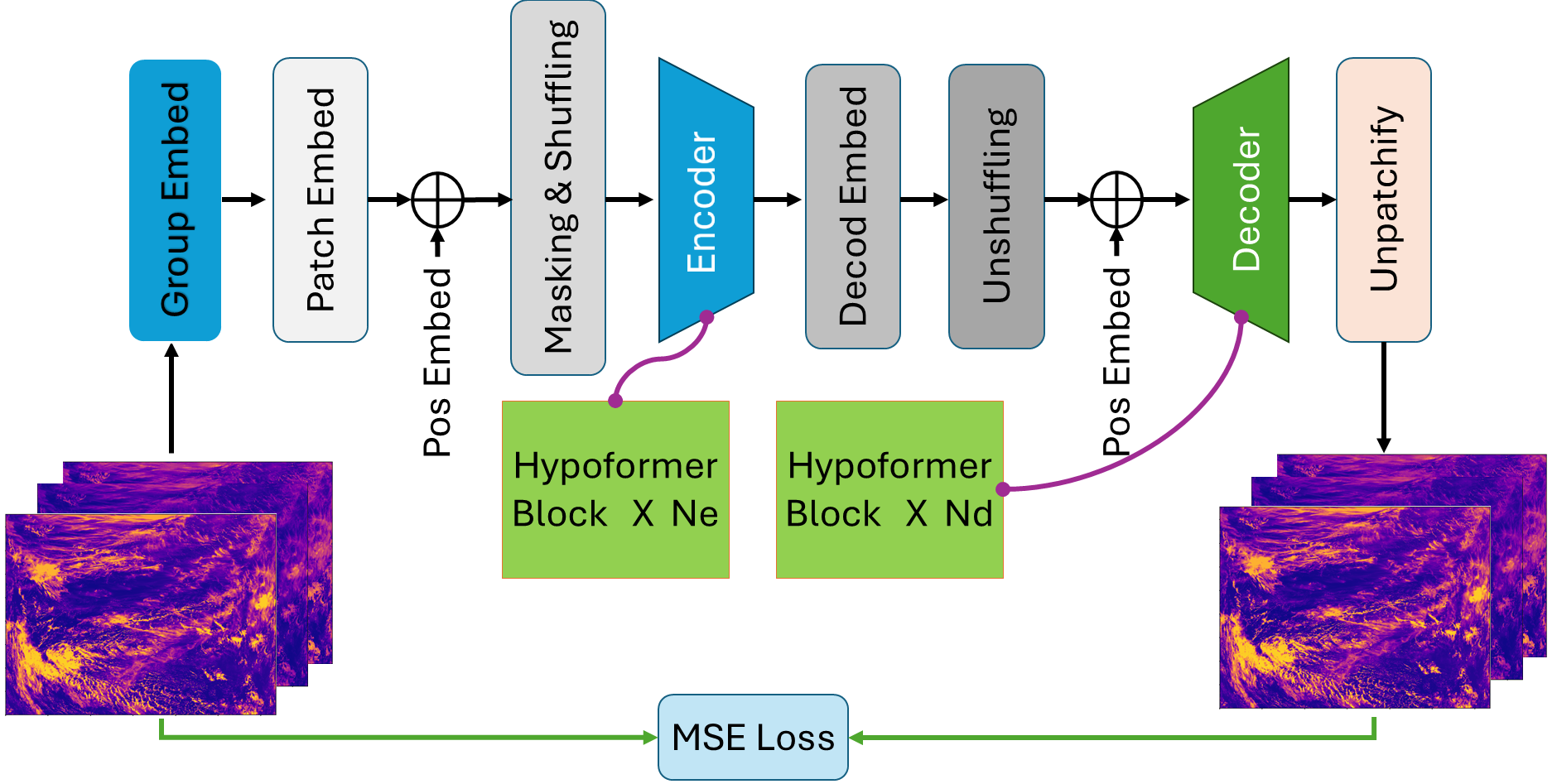}
    \caption{HyperFM architecture for Masked Autoencoder pretraining.}
    \label{fig:hyperFMg}
\end{figure}

The overall architecture of \textbf{HyperFM} is shown in \cref{fig:hyperFMg}. In our design, a \textit{\textbf{Group Embed}} module precedes the patch embedding stage and is followed by a sequence of \textbf{\textit{Hypoformer}} blocks. A central challenge in hyperspectral modeling is the large number of spectral channels. For example, our \textbf{HyperFM250K} data has 291 bands, which makes standard ViT~\cite{dosovitskiy2020image} architectures computationally expensive. Directly projecting a hyperspectral patch of size 291×8×8 into a 768 dimensional token causes severe compression loss, as the patch embedding must map a high dimensional input into a fixed token size. Even when using standard ViT blocks, the patch size must be aggressively reduced to match the token dimension, again discarding significant spectral and spatial information.

To overcome this compression bottleneck and better capture joint spectral and spatial dependencies, we introduce the \textbf{\textit{Group Embed module}}, which uses MaxViT blocks~\cite{tu2022maxvit} to process groups of spectrally adjacent channels. This module discussed below performs group-wise feature extraction and group-wise patchification before tokenization, thereby preserving substantially more spectral detail than a conventional patch embedding layer. The \textbf{\textit{Hypoformer}} blocks then use the hybrid tensor-train decomposition of~\cite{li2022hypoformer} to reduce parameter count while maintaining strong representational capacity. In our implementation, we use $N_e=4$ hypoformer blocks in the encoder and $N_d=8$ blocks in the decoder.

\begin{figure}[tb]
    \centering
    \includegraphics[width=\linewidth]{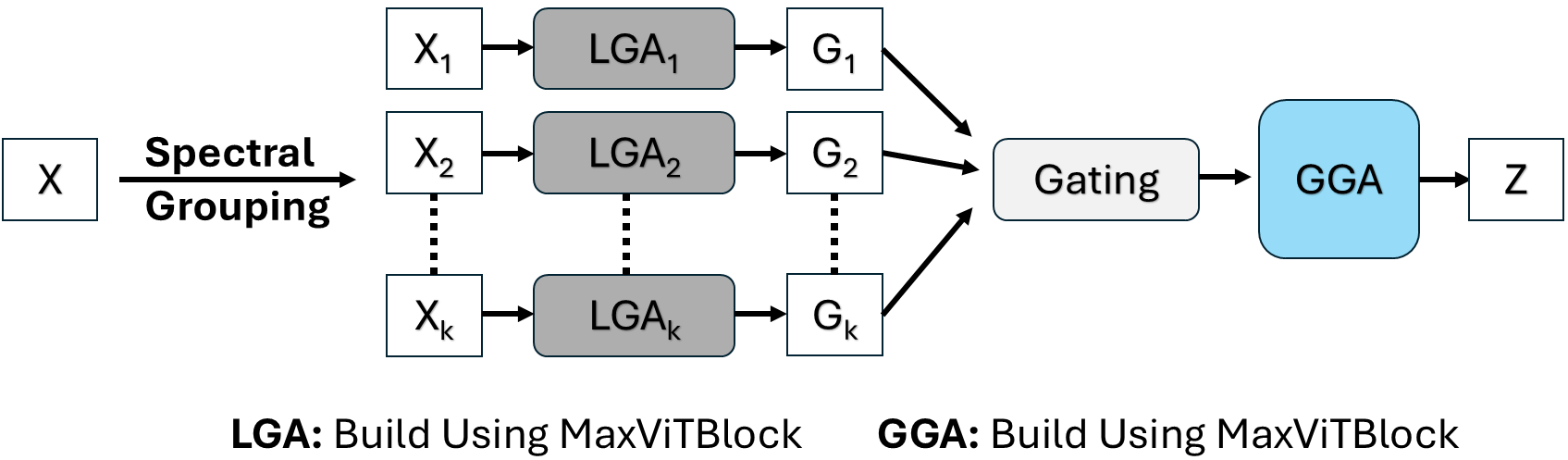}
    \caption{Group Embed module with local group attention (LGA) and global group attention (GGA).}
    \label{fig:hyperFMg2}
\end{figure}

\textbf{Group Embed Module:} 
Given a hyperspectral image \(X \in \mathbb{R}^{C \times H \times W}\), we divide the spectral dimension into \(k\) groups, producing tensors 
$
X_{1}, X_{2}, \ldots, X_{k} \in \mathbb{R}^{\frac{C}{k} \times H \times W}
$ (~\cref{fig:hyperFMg2}).
Each group is processed by an independent Local Group Attention (LGA) block, yielding intra group features 
$
G_{1}, G_{2}, \ldots, G_{k}.
$
These features are concatenated and passed into a Global Group Attention (GGA) block to model relationships across groups, producing 
$
Z \in \mathbb{R}^{C \times H \times W}.
$
Both the LGA and GGA modules are implemented using the MaxViT block~\cite{tu2022maxvit}, which applies block attention and grid attention to capture local and global spectral spatial interactions. Inspired by mixture of experts approaches~\cite{zhou2022mixture}, we introduce a gating function that selects only a subset of LGA group features before they are concatenated and passed to the GGA block. This allows the model to focus on the most informative spectral groups and improves representation efficiency. The gating parameters are fully trainable and are learned during pretraining.

For our implementation and training of our \textbf{HyperFM} model we employed spectral grouping empirically. OCI measures 119 blue bands, 163 red bands, and 9 shortwave infrared (SWIR) bands, with partial overlaps. We construct nine groups, each containing at least 13 blue bands, 18 red bands, and one SWIR band. This ensures that each group covers diverse regions of the spectrum, improving the model’s ability to learn cross spectral relationships.

\begin{figure}[tb]
    \centering
    \includegraphics[width=0.95\linewidth]{figures/hypoformer_block.png}
    \caption{Hypoformer block, which replaces standard QKV attention with HTT attention and the FFN with an LMF FFN.}
    \label{fig:hypoformer}
\end{figure}

\textbf{\textit{Hypoformer Block}}: We use the \textit{Hypoformer}~\cite{li2022hypoformer} block as the backbone of our \textbf{HyperFM} foundation model instead of a standard ViT~\cite{dosovitskiy2020image} because it provides a scalable way to reduce network parameters while preserving the representational capacity of each layer. Its hybrid architecture also allows the model to maintain full rank expressivity, unlike many other decomposition based approaches. Although \textit{Hypoformer} has demonstrated strong parameter efficiency in language models, it has not been applied to vision tasks. Here, we summarize its key components. Each Hypoformer block replaces the standard attention and feed forward layers with two modules: (1) Hybrid Tensor Train (HTT) Attention and (2) a Low Matrix Factorization (LMF) Feed Forward Network.

\textbf{HTT Attention} computes query(Q), key(K) and value (V) representations by combining two projections: a dense layer $(W_{dense})$ and a tensor train linear (TTLinear) layer $(W_{tt})$ as shown in \cref{fig:hypoformer}. Instead of producing Q, K, and V from a single matrix, HTT concatenates outputs from both layers:
{\small
\begin{equation}\label{eq.htt}
\begin{aligned}
    q_1, k_1, v_1 &= \text{Split}(X W_{\text{dense}}, 3) \\
    q_2, k_2, v_2 &= \text{Split}(X W_{\text{tt}}, 3) \\
    q &= \text{concat}(q_1, q_2) \\
    k &= \text{concat}(k_1, k_2)    \\
    v &= \text{concat}(v_1, v_2)  
\end{aligned}
\end{equation}
}

Here, $X$ is the input, $W_{dense}\in {\mathbb{R}}^{d\times 3\alpha d}$ is the dense projection layer, $W_{tt}\in {\mathbb{R}}^{d\times 3(1-\alpha)d}$ is the Tensor Train Linear layer, $d$ is the embedding dim, and $\alpha\in[0,1]$ is the compression ratio.
Once we have the q, k and v representations, we can apply the attention operation as 
${attention}(q,k,v) = \mathrm{softmax}\!\left( \frac{q k^\top}{\sqrt{d}} \right) v$

\textbf{Low Matrix Factorization Feed-Forward Network (LMF FFN)}: The LMF FFN replaces the transformer FFN with a low rank factorization based module and includes four dense layers as shown in Eq.~\ref{eq.lmf}.
{\small
\begin{equation}\label{eq.lmf}
    LMF\text{-} FFN(X) = ReLU(XU_1V_1 +b_1)U_2V_2+b_2
\end{equation}
}
where $U_1\in {\mathbb{R}}^{d\times R}$, $V_1\in {\mathbb{R}}^{R\times d_{hidden}}$, $U_2\in {\mathbb{R}}^{d_{hidden}\times R}$, $V_1\in {\mathbb{R}}^{R\times d}, b_1\in {\mathbb{R}}^{d_{hidden}}$ and $b_2\in \mathbb{R}^d$. $R$ is the matrix factorization rank.  

\paragraph{Pretraining of HyperFM:}
We pretrain HyperFM using a Masked Autoencoder (MAE) framework~\cite{reed2023scale}. As illustrated in \cref{fig:hyperFMg}, a subset of spatial patches is masked, and only the visible patches are processed by the encoder. The decoder then reconstructs the masked regions from these visible tokens. By obscuring most of the input and reconstructing only the missing content, the model is encouraged to learn global structure and long range spectral–spatial dependencies rather than rely on low level cues. This approach also avoids the complexity of contrastive or clustering based objectives, which often require large batch sizes, extensive augmentations, or sensitive hyperparameter tuning~\cite{braham2025spectralearth}. MAE further improves computational efficiency because only unmasked patches pass through the encoder, allowing us to increase encoder capacity without a proportional increase in training cost. We use an L2 reconstruction loss during pretraining.

\paragraph{Downstream Tasks:}

To evaluate our HyperFM foundation model, we select four pixel level regression tasks focused on cloud optical and microphysical property retrieval. In particular, we predict Cloud Optical Thickness (COT), Cloud Effective Radius (CER), Cloud Water Path (CWP), and Cloud Top Height (CTH), which together characterize the radiative and microphysical state of clouds. COT, also referred to as cloud optical depth, quantifies the attenuation of solar radiation as it passes through a cloud, with larger values indicating optically thicker clouds. CER describes the mean droplet size within the cloud and is directly linked to cloud albedo and precipitation processes. CWP measures the vertically integrated liquid water content and is used to estimate cloud mass and its radiative effects. CTH represents the altitude of the upper cloud boundary. These properties are fundamental to climate monitoring~\cite{ipcc_website}, weather forecasting~\cite{jones2013evaluation}, aviation safety~\cite{kessinger2008convection}, and renewable energy planning~\cite{zhou2023influences}, making accurate retrieval essential.

\begin{figure}[tb]
    \centering
    \includegraphics[width=0.99\linewidth]{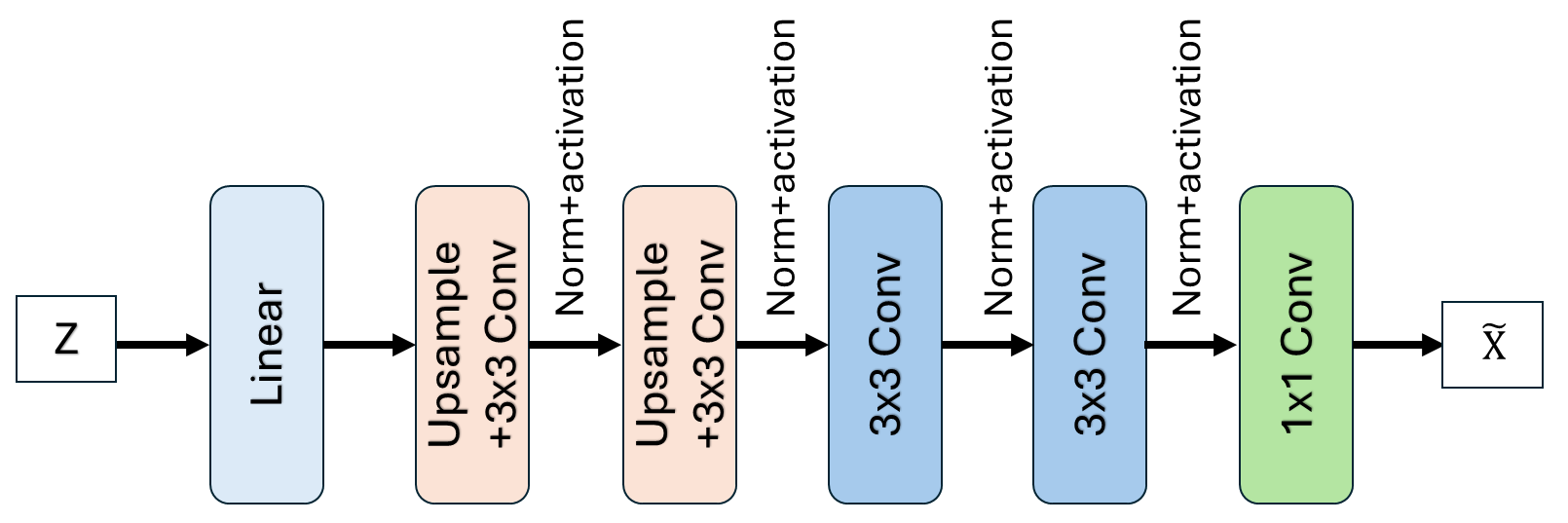}
    \caption{Lightweight Decoder for downstream evaluation.}
    \label{fig:decoder}
\end{figure}

To finetune HyperFM for these downstream tasks, we adopt a multi task learning setup, motivated by prior work showing that joint retrieval of cloud properties improves performance relative to training separate single task models~\cite{okamura2017feasibility,tushar2025joint,wang2022retrieval}. We use a lightweight decoder composed of convolutional layers, upsampling blocks, and layer normalization, as illustrated in \cref{fig:decoder}. The pretrained encoder produces a token sequence, $Z\in \mathbb{R}^{L\times d}$ which is passed to the decoder to reconstruct the target fields, $\tilde{X} \in \mathbb{R}^{4\times H \times W}$, corresponding to the four cloud property outputs.

%% file: MySections/Experiments.tex
\section{Experiments}
\label{section:experiments}
Here, we outline our experimental settings and show empirical results to address the following research questions:
(1) How well do pre-trained foundation models perform on HyperFM250K without fine-tuning?
(2) How does the proposed \textbf{HyperFM} model compare with existing hyperspectral foundation models and state-of-the-art deep learning approaches for cloud property retrievals?
(3) What are the computational costs and efficiency characteristics of HyperFM relative to existing models?  
Below, we describe our dataset, baseline comparisons, evaluation metrics, and implementation details.

\paragraph{Data:} Details of the HyperFM250K dataset are provided in~\cref{sec:data}. We will release this large scale dataset (over 3 TB) to support foundation model training and benchmarking for cloud optical and microphysical property retrieval. We also plan to extend the dataset with additional pixel level regression targets and segmentation labels, such as cloud mask generation. More details about dataset and codes are presented in the supplementary materials.

\paragraph{Baseline Models and Their Implementation:}
We compare our model \textbf{HyperFM}'s performance against both task-specific state-of-the-art models and recently proposed hyperspectral foundation models. 
\begin{itemize}
 \item \textbf{Task-specific SOTA Models: } For COT, CER, CWP, and CTH retrieval, UNet style architectures remain widely used. We include UNet~\cite{nataraja2022segmentation,li2023transfer}, CloudUNet~\cite{tushar2024cloudunet}, and CAM~\cite{tushar2025joint}. 
 These models are trained on two OCI wavelengths (660 nm, 2130 nm), following standard practice.
 \item \textbf{Hyperspectral Foundation Models:} We evaluate three recent hyperspectral foundation models: \textit{SpectralEarth}~\cite{braham2025spectralearth}, \textit{HyperSigma}~\cite{wang2025hypersigma}, and \textit{HyperFree}~\cite{li2025hyperfree}. We use the SpectralEarth ViT-B variant, the HyperSigma SpatSigma-B variant, and HyperFree ViT-B variant, all built on ViT-Base backbones. \textit{SpectralEarth} is pretrained on EnMAP imagery, while \textit{HyperSigma} is pretrained on Gaofen and EO-1 Hyperion data. \textit{HyperFree} is pretrained on Hyper-Seg~\cite{li2025hyperfree} dataset built using the AVIRIS imagery. \textit{SpectralEarth} incorporates a spectral adapter that projects inputs to 128 channels, so no spectral modification is needed; however, it requires spatial inputs of 128×128, and thus we resize \textbf{HyperFM250K} image patches from 
96×96 to 128×128. For \textit{HyperSigma}, to avoid spectral mismatch, we follow the \textit{SpectralEarth} evaluation protocol and apply linear interpolation at the patch embedding layer. We applied linear interpolation at the channel-adaptive embedding layer of \textit{HyperFree} to support patch size of 8.

\end{itemize}

\paragraph{Implementation Details:}

\textbf{HyperFM pretraining:}  
Pretraining is done on \textbf{HyperFM250K} using a Masked Autoencoder (MAE) objective. We mask 75\% of the tokens during pretraining. The model is trained for 250 epochs with early stopping based on validation MSE (patience 50). We use a batch size of 4 and the AdamW optimizer to update all parameters.

\textbf{Downstream task Finetuning:}
We evaluate all models on four downstream pixel level regression tasks: COT, CER, CWP, and CTH retrieval. For the foundation models, we freeze the pretrained encoder and attach a lightweight decoder with two convolutional layers, upsampling blocks, and layer normalization (~\cref{fig:decoder}), and only the decoder parameters are updated during finetuning. Task specific deep learning models (\textit{UNet}, \textit{CloudUNet}, \textit{CAM}) are trained using only two wavelengths, whereas foundation models (\textbf{HyperFM}, \textit{SpectralEarth}, \textit{HyperSigma}, \textit{HyperFree}) use the full 291 band spectrum. All models are trained for 250 epochs with early stopping based on validation MSE using the AdamW optimizer in a multi-task learning setup. The finetuning dataset contains 2000 training images, 250 validation images, and 2000 test images, and following~\cite{tushar2025joint}, COT and CWP are log transformed before training. OCI hyperspectral images have variable spatial dimensions, so we extract non overlapping patches of size 291×96×96 and perform inference patch wise, merging outputs to form the final retrieval. When the original spatial dimensions are not divisible by 96, we apply overlapping sampling at the borders and average the predictions in overlapping regions.

As in related work, we use Mean Squared Error (MSE) as the evaluation metric for all regression tasks.
\textit{Remarks:} \textbf{HyperFM} is pretrained on cloudy data but only on 2000 samples due to time and resource constraints. Nevertheless, it still outperforms other foundation models as shown in the following results section, demonstrating its strong potential.

%% file: MySections/Results.tex
\section{Results}
\label{section:results}

\begin{table*}[hbt]
    \centering
    \caption{Benchmark Results on Four Pixel-wise Regression Tasks [Models in \textcolor{brown}{brown} color are task-specific deep learning models]}
    \label{tab:results}
    \resizebox{0.95\linewidth}{!}{%
    \begin{tabular}{l r c c c c c r l }
        \toprule
        \textbf{Training Setup} &
        \textbf{Model Architecture} & 
        \textbf{COT MSE ($\downarrow$)} & 
        \textbf{CER MSE ($\downarrow$)} & 
        \textbf{CWP MSE ($\downarrow$)} &
        \textbf{CTH MSE ($\downarrow$)} & 
        \textbf{Model Capacity} & 
        \textbf{Trainable Param} &
        \textbf{Train Time (Hr)} \\
        \midrule
        \multirow{3}{*}{Zero-shot} 
       & SpectralEarth~\cite{braham2025spectralearth} & $61.0280$	& $14384.0316$	& $26.1292$	& $1296.5026$ & 88.78M & - & - \\ 
         & HyperSigma~\cite{wang2025hypersigma} & $53.1539$	& $39999.7969$	& $9.1567$	& $979.5114$ & 100.16M  & - & - \\ 
         & HyperFree~\cite{li2025hyperfree} & $9.170$	& $21296.765$	& $36.923$	& $1681.148$ & 101.82M & - & - \\ 
        \midrule
        \multirow{4}{*}{\shortstack{Decoder Only\\Fine-tuning}} & SpectralEarth~\cite{braham2025spectralearth} & $0.4699$	& $97.9197$	& $1.7111$	& $7.6742$ & 88.78M & 0.54M & 9.58@(3x48GB RTX8000) \\ 

          & HyperSigma~\cite{wang2025hypersigma} & $0.3212$	& $95.4874$	& $1.3317$	& $8.4936$ & 100.16M  & 0.69M & 133.31@(2x12GB 2080Ti) \\

          & HyperFree~\cite{li2025hyperfree} & $0.5570$	& $117.8959$	& $2.0632$	& $10.0716$ & 101.82M & 0.69M & 5.38@(2x48GB A6000) \\ 

          & \textcolor{blue}{\textbf{HyperFM [ours]}} & $\textcolor{blue}{\textbf{0.3124}}$	& $\textcolor{blue}{\textbf{73.7017}}$	& $\textcolor{blue}{\textbf{1.2229}}$	& $\textcolor{blue}{\textbf{5.0976}}$ & \textcolor{blue}{32.06M} & \textcolor{blue}{1.48M} & \textcolor{blue}{7.50@(2x48GB A6000)} \\
          \midrule
        
        \multirow{7}{*}{\shortstack{Full \\Fine-tuning}} &
        \textcolor{brown}{UNet}~\cite{nataraja2022segmentation,li2023transfer} & $0.3928$ & $84.7329$ & $1.5709$ & 
        $7.6786$ & 31.04M & 31.04M & 1.57@(1x48GB A6000)\\
          & \textcolor{brown}{CloudUNet}~\cite{tushar2024cloudunet} & $0.3752$	& $128.4650$ & $1.7030$	& $12.1648$ & 1.86M & 1.86M & 3.67@(1x12GB RTX2080Ti) \\
          & \textcolor{brown}{CAM}~\cite{tushar2025joint}  & $0.3367$ & $74.4473$	& $1.5123$	& $6.6306$ & 0.47M & 0.47M & 3.71@(1x12GB RTX2080Ti) \\

          & SpectralEarth~\cite{braham2025spectralearth} & $0.3404$	& $84.2870$	& $1.2545$	& $5.1727$ & 88.78M & 88.78M & 29.31@(4x24GB RTX6000) \\ 
          & HyperSigma~\cite{wang2025hypersigma} & $0.4649$	& $117.5051$	& $1.7524$	& $10.3292$ & 100.16M  & 100.16M & 4.54@(2x48GB A6000) \\
         & HyperFree~\cite{li2025hyperfree} & $0.3523$	& $86.8931$	& $1.3451$	& $6.4175$ & 101.82M & 101.82M & 3.09@(3x48GB RTX8000) \\      

          & \textcolor{blue}{\textbf{HyperFM [ours]}} & $\textcolor{blue}{\textbf{0.2615}}$	& $\textcolor{blue}{\textbf{62.3963}}$	& $\textcolor{blue}{\textbf{1.0137}}$	& $\textcolor{blue}{\textbf{4.0508}}$ & \textcolor{blue}{32.06M} & \textcolor{blue}{32.06M} & \textcolor{blue}{5.02@(2x48GB A6000)} \\ 
        \bottomrule
\end{tabular}
    }
\end{table*}

\cref{tab:results} summarizes performance across the four pixel level regression tasks. From this table, we see that \textbf{HyperFM} delivers substantial improvements over the best baseline foundation model, \textit{HyperSigma}, achieving an 18.59\% reduction in MSE for COT, 34.66\% for CER, 23.88\% for CWP, and 52.31\% for CTH. Averaged across all tasks, \textbf{HyperFM} attains a 32.36\% lower MSE compared with \textit{HyperSigma}. Notably, even the decoder only finetuned version of \textbf{HyperFM} outperforms all competing hyperspectral foundation models as well as task-specific state-of-the-art architectures, demonstrating the strength of our pretrained \textbf{HyperFM} encoder. We also note that the existing foundation models (HyperFree, HyperSigma and SpectralEarth) perform poorly without finetuning, since they were pretrained on cloud-free hyperspectral data.

\begin{figure*}
    \centering
    \includegraphics[width=0.99\linewidth]{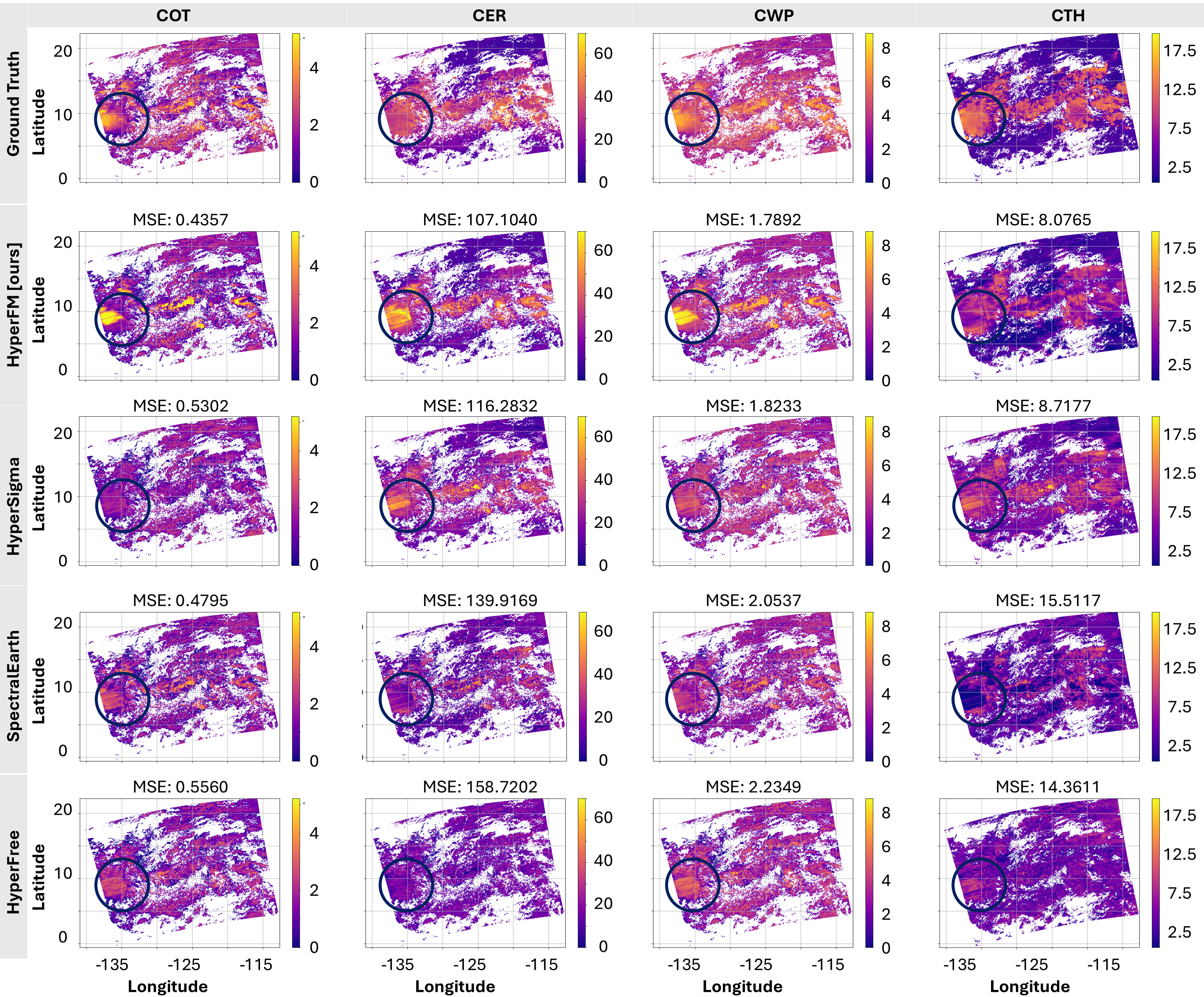}
    \caption{Comparison of Hyperspectral FMs on four pixel-wise regression tasks: cloud optical thickness(COT), cloud effective radius (CER), cloud water path (CWP) and cloud top height (CTH). MSE score for each retrievals are reported on top of the retrieval image. Sometimes grids are noticeable at the patch borders which reduces with overlapped sampling at additional computational cost.}
    \label{fig:result}
\end{figure*}

~\cref{fig:result} provides qualitative comparisons among \textit{SpectralEarth}, \textit{HyperSigma}, and our \textbf{HyperFM} on the COT, CER, CWP, and CTH retrieval tasks, with the corresponding MSE values reported above each prediction. For a fair comparison, we use only the decoder only finetuned variant of HyperFM (HyperFM-dec-only-FT), consistent with the finetuning setup of all other foundation models. Across all retrievals, \textbf{HyperFM} produces visibly more accurate spatial patterns and yields lower reconstruction errors than the baselines. White regions in the figure denote invalid pixels where official PACE OCI products or ground truth labels are unavailable; these areas are omitted from all quantitative evaluations. We observe that \textit{HyperSigma} and \textit{SpectralEarth} systematically underestimate key cloud properties, while HyperFM exhibits mild overestimation in isolated regions yet captures global patterns more faithfully. \textit{SpectralEarth} fails to generate predictions in certain areas, particularly for CER and CTH, likely due to its pretraining on cloud free land scenes. Both \textit{SpectralEarth} and \textit{HyperSigma} are pretrained exclusively on cloud free hyperspectral datasets, which limits their ability to model cloud radiative and microphysical variability and explains their comparatively weaker downstream performance.

%% file: MySections/Discussions.tex
\subsection{Discussions}
Since HyperFM relies on Hybrid Tensor Train (HTT) operations, we report its computational complexity here. The time complexity of HTT is
$
\mathcal{O}\!\left(\alpha N^{2} + D\,\big(\max[\alpha N,(1-\alpha)N]\big)^{1+\frac{1}{D}} R^{2}\right),
$
where $\alpha$ is the compression ratio $([0,1])$, $N$ is the original matrix dimension, $D$ is the number of TT cores, and $R$ is the TT rank~\cite{li2022hypoformer}. In our implementation, we set $\alpha = 0.50$, use $D = 3$ TT cores, and specify the TT rank as $R = 3$. The corresponding space complexity is
$
\mathcal{O}\!\left(\alpha N^{2} + D(1-\alpha)^{\frac{1}{D}} N^{\frac{2}{D}} R^{2}\right).
$

\textbf{Interpretation:} HTT replaces a large dense projection with a compact tensor train factorization, reducing both computational and memory costs while maintaining full-rank expressivity. This makes HTT significantly more efficient than the standard ViT projection layer, which incurs $\mathcal{O}(N^{2})$ complexity and scales poorly for high spectral dimensionality. In contrast, HTT exhibits subquadratic growth in practice due to the factorized term $N^{1+\frac{1}{D}}$ and the small TT rank $R$, enabling \textbf{HyperFM} to process high-dimensional hyperspectral inputs more effectively and with far fewer parameters than conventional ViT architectures.

\textbf{Retrieval Quality:}
Current retrieval quality (~\cref{fig:result}) is not yet optimal, particularly near spatial borders where patch discontinuities introduce artifacts. Using overlapping patches during inference may help smooth transitions and reduce boundary errors.

\textbf{Limitations:} 
HyperFM currently uses a fixed configuration with a 50\% compression ratio, three TT cores, and TT rank of three. A detailed ablation study on these hyperparameters is planned. Ground truth labels are derived from PACE‐OCI Level-2 products produced by optimal inversion algorithms, which may introduce statistical biases. To obtain more accurate supervision, we intend to co-locate PACE radiances with active sensor measurements once they become available; at the time of this writing, no collocated active sensor data exist for the four cloud properties examined.
Finally, model benchmarking was performed on a subset of the full dataset due to computational constraints. We are pursuing collaborations to expand compute capacity for large-scale experiments. Despite these limitations, HyperFM consistently outperforms existing hyperspectral foundation models and task-specific baselines, demonstrating strong generalization across all evaluated pixel-level regression tasks.

%% file: MySections/Conclusions.tex
\section{Conclusions}
Current hyperspectral foundation models are pretrained on cloud-free datasets, limiting their applicability to atmospheric retrieval tasks. We address this gap by introducing \textbf{HyperFM250K}, a large PACE-OCI hyperspectral dataset with over 60\% cloud coverage and four pixel-level cloud property retrieval tasks. We also propose \textbf{HyperFM}, a parameter-efficient foundation model designed to learn robust representations from large-scale cloudy hyperspectral data. Across all downstream cloud retrieval tasks, \textbf{HyperFM} outperforms existing hyperspectral foundation models and task-specific baselines. In future work, we will expand \textbf{HyperFM250K} with additional tasks and evaluate \textbf{HyperFM} on broader HSI benchmarks to further test its generalization capabilities.

%% file: MySections/ack.tex
\section*{Acknowledgment}
This research is partially supported by grants from NSF 2238743, and UMBC COEIT Summer Student Project Award.

This work was carried out using the computational facilities of the High Performance Computing Facility, University of Maryland Baltimore County. - https://hpcf.umbc.edu/

%% file: MySections/suppl.tex
\clearpage
\setcounter{page}{1}
\maketitlesupplementary

\section{Our HyperFM250k Dataset}
\label{sec:data_suppl}
\begin{figure*}[t]
    \centering
    \includegraphics[width=0.99\linewidth]{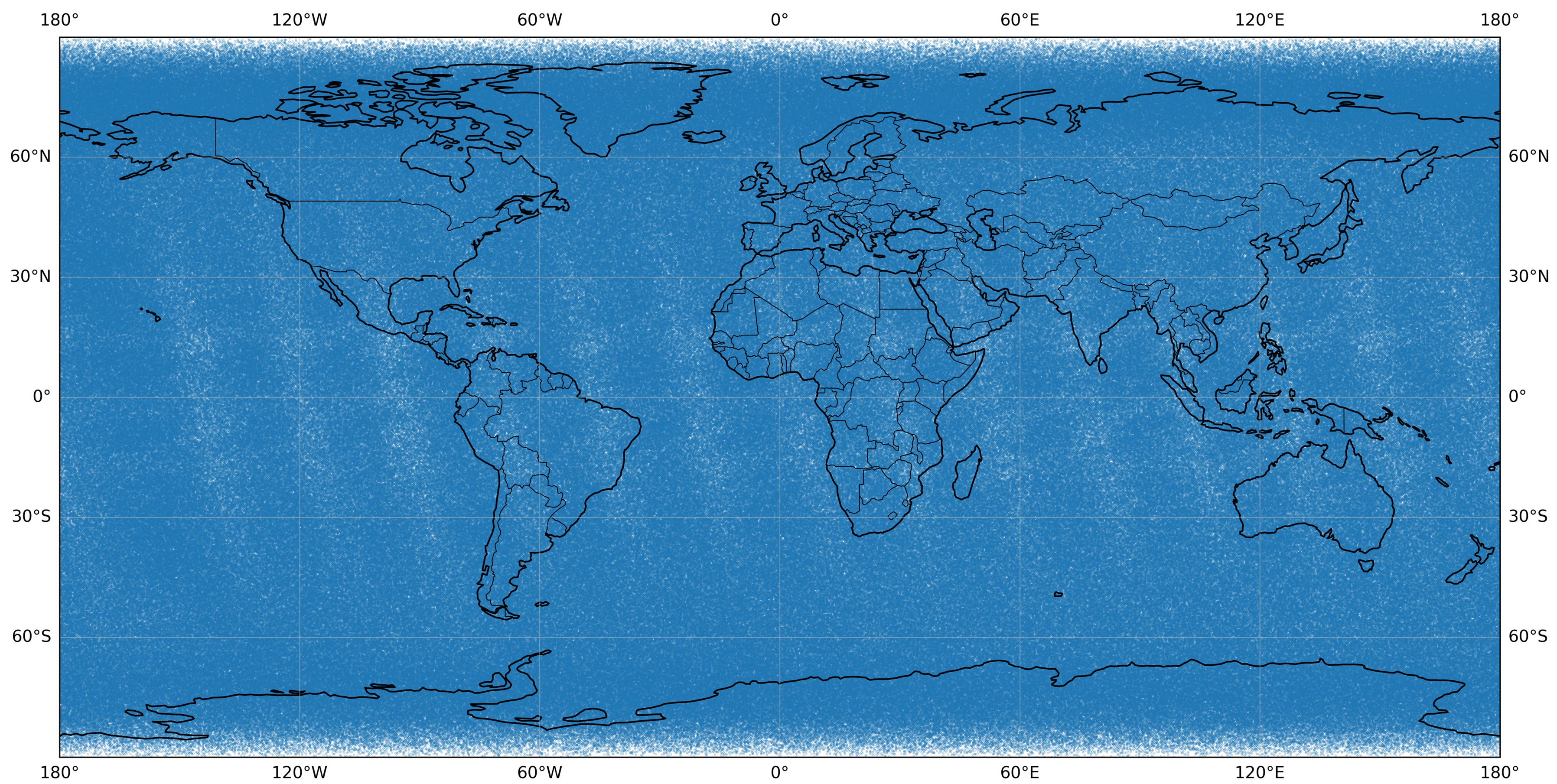}
    \caption{Geographical scatter plot of \textbf{HyperFM250k}. The map depicts the global
coverage of hyperspectral images within our dataset, demonstrating its extensive
geographical scope.}
    \label{fig:coverage}
\end{figure*}

Hyperspectral imaging from space offers detailed spectral information about the Earth’s surface and atmosphere, and recent missions have significantly increased the volume and quality of available data. Notable examples include EnMAP~\cite{kaufmann2012science}, PRISMA~\cite{pignatti2013prisma}, and the forthcoming CHIME mission~\cite{nieke2018towards}. These systems are optimized for land-focused applications and provide the spatial, spectral, and radiometric performance required for tasks such as vegetation characterization, mineralogical mapping, and terrestrial environmental monitoring. Their design, however, limits their suitability for marine, cryospheric, and atmospheric observations. Over ocean and polar regions, the signal is weaker and more sensitive to calibration uncertainties, while atmospheric studies often require broader spectral coverage, higher radiometric stability, and measurement strategies tailored to cloud and aerosol retrievals. Consequently, although current missions substantially advance land-oriented hyperspectral remote sensing, they provide only partial support for applications central to climate research and numerical weather prediction, underscoring the need for instruments explicitly designed for these domains.

NASA’s PACE mission directly addresses the limitations of land-oriented hyperspectral systems by introducing the Ocean Color Instrument (OCI), which delivers global hyperspectral coverage tailored to oceanic and atmospheric applications. OCI provides continuous spectral sampling from roughly 340 to 890 nm at 5 nm resolution, along with seven additional shortwave infrared channels centered at 940, 1038, 1250, 1378, 1615, 2130, and 2260 nm. This combination offers the spectral density and calibration stability required for robust retrievals of ocean color, cloud properties, and aerosol composition. In contrast, earlier multispectral instruments such as MODIS~\cite{platnick2015modis} and VIIRS~\cite{platnick2019viirs} have supported long-term atmospheric and ocean observations but lack the spectral detail and radiometric performance needed for reliable microphysical characterization. As a result, PACE OCI functions as a complementary capability to existing hyperspectral missions while providing the spectral depth necessary for improved environmental and climate research.

\subsection{Dataset Overview}
\textbf{HyperFM250k} is a hyperspectral dataset built from PACE-OCI observations. It contains 250k HSI tiles with cloud coverage more than 60\%, addressing the limitations of earlier releases that often kept the cloud presence below ten percent.

\subsubsection{Raw Data Overview}
A summary of the raw PACE-OCI data used to generate \textbf{HyperFM250k} is shown in \cref{tab:data_sum_sup}. PACE-OCI provides Level-1 radiance data in three forms: A) raw, B) calibrated and geolocated, and C) calibrated, geolocated, and co-registered to a common grid. Because the raw measurements require additional calibration and are not reliable for direct analysis, Level-1B and Level-1C are the practical options. However, level 1C has a coarse spatial resolution of 5km while level 1B offers finer resolution of 1.2km. Therefore we select level 1B observations to construct our dataset. An additional advantage of Level-1B is its colocation with Level-2 products which simplifies downstream use without requiring further preprocessing and alignment. Note that all geophysical science products are released as level-2 products. Some of the key geophysical products include Cloud Optical Thickness (COT), Cloud Effective Radius (CER), Cloud Water Path (CWP), and Cloud Top Height (CTH), among others~\cite{nasa_pace_v3}.

A total of 2262 granules were downloaded using a python script specifying the search criterion. The granules were selected from May 2024 to April 2025. The \textit{train} split comprises 12 days of data from 12 different months, while the validation and test splits each use 2 days. The level-1B raw granules totaled approximately $3,625\ \text{GB}$ while level-2 granules were approximately $181\ \text{GB}$.




        
        
        
        




\begin{table*}[hbt]
\centering
\caption{Summary of PACE OCI Data used to construct \textbf{HyperFM250k} dataset}
\label{tab:data_sum_sup}

\resizebox{0.95\linewidth}{!}{
\begin{tabular}{lccccc}
\toprule
\textbf{Product} &
\textbf{Processing Level} &
\textbf{Variables Used} &
\textbf{Spectral Bands} &
\textbf{Temporal Coverage} &
\textbf{\# Raw Granules} \\
\midrule

PACE\_OCI\_L1B\_SCI &
Level-1B &
\makecell[c]{observation\_data: \\
rhot\_blue, qual\_blue \\
rhot\_red, qual\_red \\
rhot\_SWIR, qual\_SWIR} &
\makecell[c]{blue (\#119): 314--605 nm \\
red (\#163): 600--895 nm \\
SWIR (\#9): 940--2258 nm \\
Total: \#291} &
\makecell[c]{Train: 2024-05-10, 2024-06-10, 2024-07-10 \\
2024-08-10, 2024-09-11, 2024-10-10 \\
2024-11-10, 2024-12-10, 2025-01-10 \\
2025-02-10, 2025-03-10, 2025-04-10 \\
Valid: 2025-02-28, 2025-04-28 \\
Test: 2025-02-20, 2025-03-20} &
\makecell[c]{Train: 1677 \\
Valid: 291 \\
Test: 294} \\

\midrule

PACE\_OCI\_L2\_CLOUD &
Level-2 &
\makecell[c]{geophysical\_data: \\
cot\_21, cer\_21 \\
cwp, cth} &
Same as above &
Same as above &
\makecell[c]{Train: 1687 \\
Valid: 291 \\
Test: 294} \\

PACE\_OCI\_L2\_CLOUD\_MASK &
Level-2 &
\makecell[c]{geophysical\_data: \\
cloud\_flag \\
cloud\_flag\_dilated} &
Same as above &
Same as above &
\makecell[c]{Train: 1687 \\
Valid: 291 \\
Test: 294} \\

\bottomrule
\end{tabular}
}
\end{table*}

\subsubsection{Preprocessed Dataset Overview}
Both Level 1B and Level-2 raw granules are preprocessed using the \cref{alg:one}. Preprocessed hyperspectral tiles were saved in (.npy) format. A summary of the HyperFM250k is as follow:

\begin{itemize}
    \item 253,104 hyperspectral tiles. Each tile has input and corresponding target data.
    \item Train / Val / Test: 168,394 /42,355 /42,355
    \item Train / Val / Test: 1688.7 GB / 431.73 GB / 431.73 GB
    \item Each Tile: NumPy Array shaped: $291\times 96\times 96$
    \item Per-channel mean and standard deviation provided
\end{itemize}

\subsection{Tasks}
The \textbf{HyperFM250k} dataset includes four pixel-wise regression tasks and one binary segmentation task.

\begin{itemize}
    \item Cloud Optical Thickness (COT) – Also called cloud optical depth. It measures how much incoming solar radiation is attenuated as it passes through a cloud, with larger values indicating optically thicker clouds.
    \item Cloud Effective Radius (CER) – Describes the mean droplet size within the cloud and is closely linked to cloud albedo and precipitation processes.
    \item Cloud Water Path (CWP) – Represents the vertically integrated liquid water content, providing an estimate of total cloud water and its radiative influence.
    \item Cloud Top Height (CTH) – Indicates the altitude of the cloud’s upper boundary.
    \item Cloud Mask (binary segmentation) – Identifies whether each pixel is cloud or non-cloud, supporting cloud screening and downstream retrieval quality.
\end{itemize}

These five tasks capture core aspects of cloud structure and behavior, and they play a central role in climate monitoring~\cite{ipcc_website}, weather forecasting~\cite{jones2013evaluation}, aviation safety~\cite{kessinger2008convection}, and renewable energy studies~\cite{zhou2023influences}.

Due to time and space limitations, we were only able to report on the four regression tasks. 

\begin{figure*}[h]
    \centering
    \includegraphics[width=0.99\linewidth]{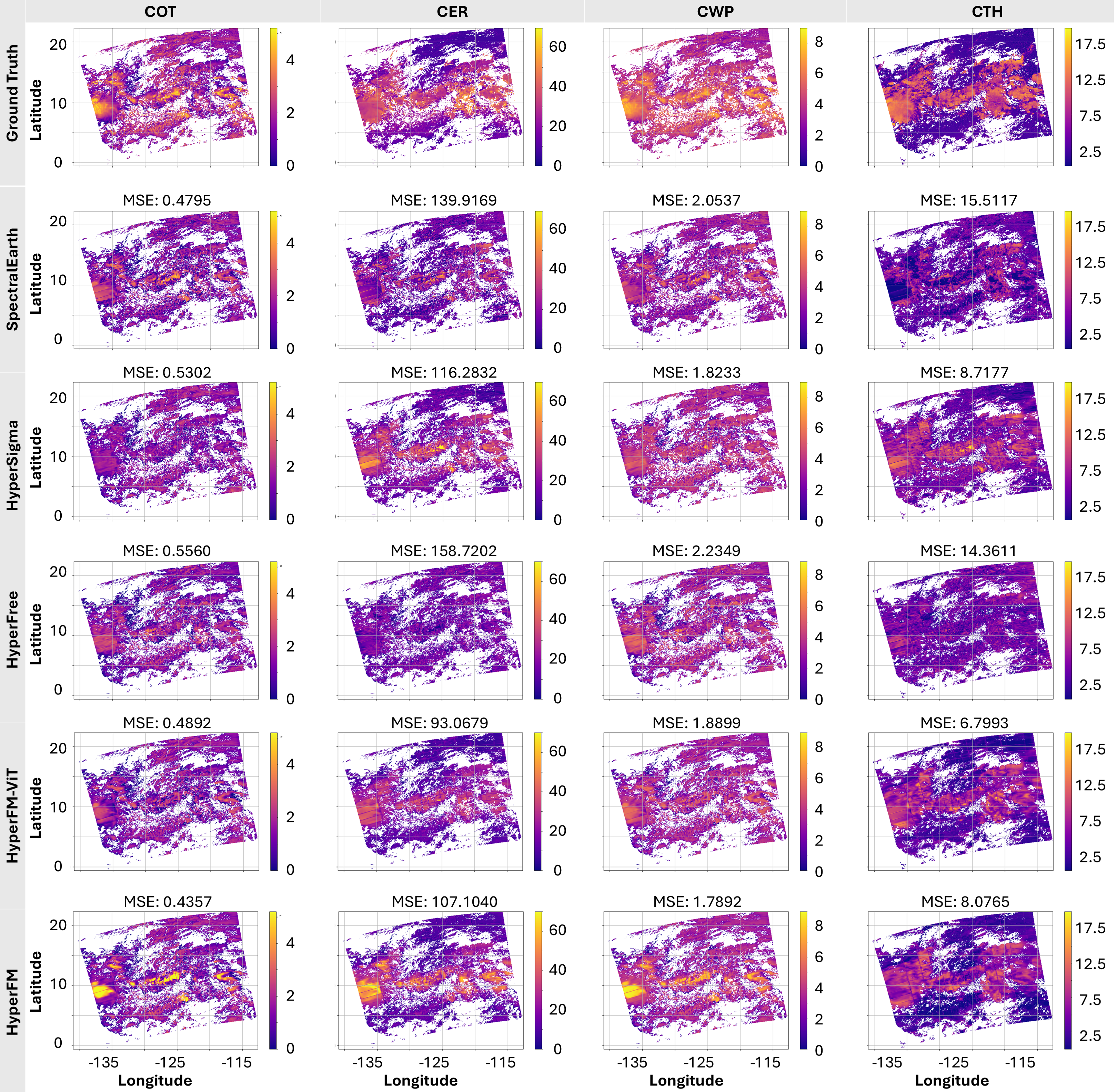}
    \caption{Comparison of Hyperspectral FMs on four pixel-wise regression tasks: cloud optical thickness(COT), cloud effective radius (CER), cloud water path (CWP) and cloud top height (CTH). MSE score for each retrievals are reported on top of the retrieval image. Sometimes grids are noticeable at the patch borders which reduces with overlapped sampling at additional computational cost.}
    \label{fig:suppl_results}
\end{figure*}

\begin{table*}[hbt]
    \centering
    \caption{Additional Benchmark Results on Four Pixel-wise Regression Tasks}
    \label{tab:add_results}
    \resizebox{0.95\linewidth}{!}{%
    \begin{tabular}{r c c c c c c r l }
        \toprule
        \textbf{Model Architecture} & 
        \textbf{COT MSE ($\downarrow$)} & 
        \textbf{CER MSE ($\downarrow$)} & 
        \textbf{CWP MSE ($\downarrow$)} &
        \textbf{CTH MSE ($\downarrow$)} & 
        \textbf{Model Capacity} & 
        \textbf{Trainable Param} &
        \textbf{Train Time (Hr)} \\
        \midrule
        UNet~\cite{nataraja2022segmentation,li2023transfer} & $0.3928$ & $84.7329$ & $1.5709$ & 
        $7.6786$ & 31.04M & 31.04M & 1.57@(1x48GB A6000)\\
        CloudUNet~\cite{tushar2024cloudunet} & $0.3752$	& $128.4650$ & $1.7030$	& $12.1648$ & 0.1.86M & 0.1.86M & 3.67(@1x12GB RTX2080Ti) \\
        CAM~\cite{tushar2025joint}  & $0.3367$ & $74.4473$	& $1.5123$	& $6.6306$ & 0.47M & 0.47M & 3.71(@1x12GB RTX2080Ti) \\
        \midrule
        SpectralEarth~\cite{braham2025spectralearth}[w/o FT] & $61.0280$	& $14384.0316$	& $26.1292$	& $1296.5026$ & 88.78M & - & - \\ 
        HyperSigma~\cite{wang2025hypersigma}[w/o FT] & $53.1539$	& $39999.7969$	& $9.1567$	& $979.5114$ & 100.16M  & - & - \\ 
        HyperFree~\cite{li2025hyperfree}[w/o FT] & $9.170$	& $21296.765$	& $36.923$	& $1681.148$ & 101.82M & - & - \\ 
        \midrule
        SpectralEarth~\cite{braham2025spectralearth} & $0.4699$	& $97.9197$	& $1.7111$	& $7.6742$ & 88.78M & 0.54M & 9.58@(3x48GB RTX8000) \\ 
        HyperSigma~\cite{wang2025hypersigma} & $0.3212$	& $95.4874$	& $1.3317$	& $8.4936$ & 100.16M  & 0.69M & 133.31@(2x12GB 2080Ti) \\

        HyperFree~\cite{li2025hyperfree} & $0.5570$	& $117.8959$	& $2.0632$	& $10.0716$ & 101.82M & 0.69M & 5.38@(2x48GB A6000) \\ 
        \midrule
        HyperFM-ViT-Dec Only FT & $0.3345$	& $74.2573$	& $1.2792$	& $5.4212$ & 49.54M & 1.48M & 6.07@(3x48GB RTX8000) \\ 
        HyperFM-ViT-Full FT & $0.2672$	& $59.3235$	& $1.0384$	& $4.2518$ & 49.54M  & 49.54M & 3.53@(3x48GB RTX8000) \\
        \textbf{HyperFM-Dec Only FT [ours]} & $\textbf{0.3124}$	& $\textbf{73.7017}$	& $\textbf{1.2229}$	& $\textbf{5.0976}$ & 32.06M & 1.48M & 7.50@(2x48GB A6000) \\
        \textbf{HyperFM-Full FT [ours]} & $\textbf{0.2615}$	& $\textbf{62.3963}$	& $\textbf{1.0137}$	& $\textbf{4.0508}$ & 32.06M & 32.06M & 5.02@(2x48GB A6000) \\ 
        \bottomrule
    \end{tabular}
    }
\end{table*}

\section{Additional Results}
We present additional results from \cref{section:results} here which were excluded due to space limitation. We compared with another recent hyperspectral foundation model called HyperFree~\cite{li2025hyperfree} by loading their ViT-base weights and adding the convolutional decoder as shown in \cref{fig:decoder}. Note that we remove the \textit{neck} from HyperFree ViT-b encoder for fair comparison with all the HFMs. The model was finetuned in the same way the other HFMs [see ~\cref{section:experiments}] were, that is for 250 epochs with an early stopping criterion set to validation MSE loss and a patience of 30. Furthermore, we replaced the Hypoformer block in our HyperFM with regular ViT blocks and called it HyperFM-ViT. We did this to see the trade-off between using \textit{hypoformer} and ViT blocks. HyperFM-ViT are finetuned both decoder only and full network with the same hyperparameters. All the results are reported in \cref{tab:add_results}. We can see that our \textit{HyperFM} outperforms \textit{HyperFree} along with other variants of HyperFM. The results entail that ViT blocks also achieve similar results but with additional computational cost. It also entails that the key advantage of our model stems from the \textit{Group Embed} module while \textit{Hypoformer} blocks keep it parameter efficient.

\cref{fig:suppl_results} provides a qualitative comparison of all the SOTA HFMs: SpectralEarth~\cite{braham2025spectralearth}, HyperSigma~\cite{wang2025hypersigma}, HyperFree~\cite{li2025hyperfree}, HyperFM-ViT, and \textbf{HyperFM}. For fairness, all retrieval examples are drawn from the decoder-only fine-tuned models. \cref{section:results} discusses the SpectralEarth and HyperSigma results with respect to our HyperFM. Beside that we can see that HyperFree shows clear difficulty in recovering the CER property same as \textit{SpectralEarth} and consistently falls behind other hyperspectral foundation models, including SpectralEarth, HyperSigma, and HyperFM in COT, CWP and CTH retrievals. Its design choice to suppress cloudy pixels further limits its ability to handle cloud-centric tasks, which explains the performance gap observed here.